# Using fixed and mobile eye tracking to understand how visitors view art in a museum: A study at the Bowes Museum, County Durham, UK

*Claire Warwick (c.l.h.warwick@durham.ac.uk), Durham University, United Kingdom*

*Andrew Beresford (a.m.beresford@durham.ac.uk), Durham University, United Kingdom*

*Soazig Casteau (soazig.casteau@durham.ac.uk), Durham University, United Kingdom*

*Hubert P. H. Shum (hubert.shum@durham.ac.uk), Durham University, United Kingdom*

*Dan Smith (daniel.smith2@durham.ac.uk), Durham University, United Kingdom*

*Francis Xiatian Zhang (xiatian.zhang@durham.ac.uk), Durham University, United Kingdom*

The following paper describes a collaborative project involving researchers at Durham University, and professionals at the Bowes Museum, Barnard Castle, County Durham, UK, during which we used fixed and mobile eye tracking to understand how visitors view art.

Our study took place during summer 2024 and builds on work presented at DH2017 (Bailey-Ross *et al.*, 2017). Our interdisciplinary team included researchers from digital humanities, psychology, art history and computer science, working in collaboration with professionals from the museum. We used fixed and mobile eye tracking to understand how museum visitors view art in a physical gallery setting. This research will enable us to make recommendations about how the Museum's collections could be more effectively displayed, encouraging visitors to engage with them more fully.

Research context

The Bowes Museum

Now under new management, the Bowes Museum is going through an unprecedented transformation. It has restructured its business strategy, temporary exhibition programme, and outreach and public engagement policies. But it is unhappy with the way that its permanent collections are displayed, including captions, wall text and colour, contextualizing information and lighting. Its curators also do not know how visitors negotiate gallery spaces, or how they view art. They therefore commissioned our study. This work builds on more than a decade of collaborative engagement between Durham University and museums in County Durham, focused on raising the profile of the collections of County Durham and collaborating with our communities.

Previous studies of eye tracking and art

Our experience of art is a product of the interaction of several cognitive and affective processes, the first of which is a visual scan. When viewing an artwork, observers gather information through a series of fixations, interspersed by rapid eye movements, known as saccades. The direction of saccades is determined by an interaction between the goals of the observer and the physical properties of the scene such as colour, texture and brightness. Studying eye movements offers an insight, based on quantitative data, that does not depend on the beliefs, memories, or subjective impressions of participants. It has been widely used in Human Computer Interaction studies, to complement qualitative methods such as think-aloud protocols (Bergstrom and Schall, 2014).

Previous eye-tracking research on art has highlighted its potential to transform how we understand visual processing in the arts (Bindemann *et al.*, 2005; Massaro *et al.*, 2012; Brieber *et al.*, 2014) while also offering a direct way of studying museum/gallery visits (Heidenreich and Turano, 2011; Walker *et*

*al.*, 2017). Most eye-tracking studies have been conducted in the laboratory, using images of paintings on a digital screen. This setting ensures control over properties such as image size, colour and light. However, the user experience differs significantly from a gallery/museum visit.

Several studies show that context can influence the aesthetic experience of artworks (Brieber *et al.*, 2014; Blandford, Furniss and Makri, 2016; Carbon, 2017). Rogers, for example, shows how people in a museum or gallery come to understand and appropriate technologies in their own terms and for their own situated purposes (Rogers, 2012). Studies of the link between art and aesthetic pleasure suggest that viewers may enjoy art because it makes them feel happy, or because acquiring information about the artwork gives them intellectual satisfaction. Thus, a viewer may be pleased to learn that a painting is from Picasso's blue period, even if its subject feels intrinsically melancholy (Leder, Carbon and Ripsas, 2006; Melcher and Bacci, 2013).

While eye-tracking data offer rich insights into art viewing, analysis of its results often demands significant human effort, particularly in identifying Areas-of-Interest (AOIs) from participants' gaze patterns (Jongerius *et al.*, 2021). Recent advances in deep learning systems, however, leverage artificial neural networks to detect and classify complex data patterns, allowing gaze video stream annotation to be automated (LeCun, Bengio and Hinton, 2015; Deane, Toth and Yeo, 2022; Trajkovska, Kljun and Pucihar, 2024). However, to the best of our knowledge, no existing work has combined automatic gaze AOI recognition with eye-tracking studies of viewing art in real-world gallery scenarios.

Methods

To understand how visitors negotiate gallery spaces, what they look at and how long they look at objects, we used mobile eye tracking to capture data on the movements of 125 volunteers throughout the Museum. We then used computer modelling to analyse the resulting videos to determine what visitors looked at, whether this was different from what they reported looking at and whether they tended to fixate on faces. We used a fine-tuned YOLOv8 model (Varghese and Sambath, 2024) to detect whether a painting overlaps with the gaze focus point and employed the RetinaFace model (Deng *et al.*, 2019) to determine whether a face within the painting aligns with the gaze focus point. A time counter was applied to automatically generate statistics on how long participants focused on each painting and specific human features (e.g., faces) within the paintings.

Our previous research, which used digital copies of paintings housed at Aukland Castle, County Durham, demonstrated that semantic context provided by descriptive labels is an important factor in the enjoyment of digital artworks. If the content of labels changes, visitors look at different parts of the painting, but if this causes them to look away from faces, they are less likely to enjoy the painting(Bailey-Ross *et al.*, 2019). But is this true of real artworks? To understand this, we conducted a second experiment using fixed eye-tracking, capturing the responses of 50 volunteers to test the usefulness of existing contextualizing information such as gallery labels or wall text relevant to the entire room. We investigated whether such text could be rewritten to improve visitors' aesthetic experience and cognitive understanding.

Experimental data was then cross-referenced with questionnaires. These captured basic demographics and used Likert scales to record psychological responses, including information about whether the viewers enjoyed the art, what they remembered seeing and whether they experienced any difficulties due to lighting or the way that pictures were hung.

Findings

Data is still being analysed, however initial findings suggest that although most visitors enjoyed the artworks, they found it difficult to see some paintings due to poor lighting and the height at which they were hung. Some visitors did not recall seeing some of the most important paintings in the gallery. But visitor experience was improved by changes in labelling. Once analysis is completed, our findings will allow us to make detailed recommendations about how the presentation of the art could be improved to support visitor engagement.